\documentclass[11pt,a4paper]{article}
\usepackage[hyperref]{acl2020}
\usepackage{times}
\usepackage{latexsym}

\usepackage{microtype}

\usepackage{times}
\usepackage{latexsym}
\usepackage{natbib}

\usepackage{amsmath, amsfonts, comment}
\usepackage{graphicx}
\usepackage{algorithm, algpseudocode}

\usepackage{url}

\def\R{\mathbb{R}}
\def \1{\textrm{1}}

\aclfinalcopy 
\def\aclpaperid{208} 

\title{Learning Robust Models for e-Commerce Product Search}

\author{Thanh V. Nguyen \\
  Iowa State University \\
  \texttt{thanhng@iastate.edu} \\\And
  Nikhil Rao \\
  Amazon \\
  \texttt{nikhilsr@amazon.com} \\\And
  Karthik Subbian  \\
  Amazon \\
  \texttt{ksubbian@amazon.com} \\
  }

\date{}

\begin{document}
\maketitle
\begin{abstract}
  Showing items that do not match search query intent degrades customer experience in e-commerce. These mismatches result from counterfactual biases of the ranking algorithms toward noisy behavioral signals such as clicks and purchases in the search logs. Mitigating the problem requires a large labeled dataset, which is expensive and time-consuming to obtain.  In this paper, we develop a deep, end-to-end model that learns to effectively classify mismatches and to generate hard mismatched examples to improve the classifier. We train the model end-to-end by introducing a latent variable into the cross-entropy loss that alternates between using the real and generated samples. This not only makes the classifier more robust but also  boosts the overall ranking performance. Our model achieves a relative gain compared to baselines by over $26\%$ in F-score, and over $17 \%$ in Area Under PR curve. On live search traffic, our model gains significant improvement in multiple countries. 
\end{abstract}

\section{Introduction}
\label{sec:intro}
Deep learning models have shown excellent performance in the natural language domain, and this success has inspired practitioners to adapt these models to information retrieval tasks \cite{mitra2017learning, huang2013learning}. However, deep learning has not succeeded in these tasks due to the lack of massive labeled datasets \citep{dehghani2017neural}. 
Another reason is that word-based representations \citep{mikolov2013distributed, pennington2014glove} are less useful in representing complex, informal search queries \cite{xiong2017end} and hence provide limited understanding of the search intent. 
In the absence of explicit knowledge of which documents are ``matched" with a search query and which are ``mismatched", it is hard to learn robust deep learning models that understand the query intent and  find high-quality, relevant documents. 

Text-based product search is even more challenging. Simple modifications to the input query (or a product title) can completely change the search intent (or the product type, respectively). Take, for example, the query \texttt{gray iPhone X} by which a user is looking for a specific phone. Slightly modified queries such as \texttt{iPhone X charger} and \texttt{case for iPhone X} refer to different products. Therefore, it is hard for distributed representations to capture the nuances. Moreover, noisy user-behavioral signals from clicks and purchases (e.g., users purchased a phone while searching for a charger) can lead to biases in the ranking algorithms. As such, even top-ranked items may not match the search intent. 

\begin{figure}
\centering
\includegraphics[width = 60mm, height= 40mm]{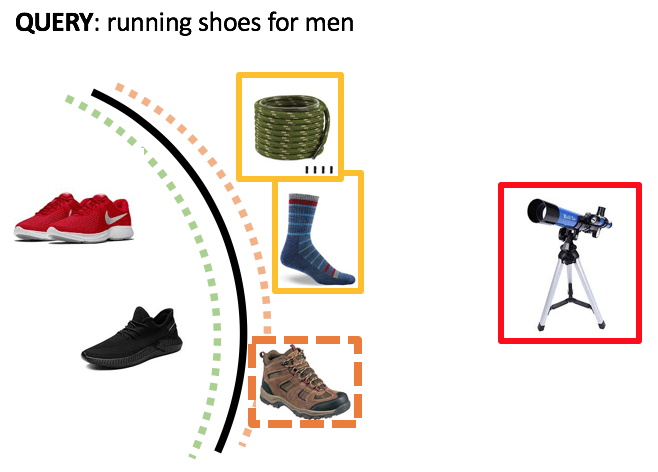}
\caption{\small (Best seen in color) The query is \texttt{running shoes for men}. The solid black line illustrates the classification boundary, and each dotted line is a small margin around the boundary. Samples to the left of the boundary are matched. To the right are mismatches, such as socks and shoelaces (orange boxes). To the extreme right is a telescope, which is an easy-to-classify example (red box). Close to the classification boundary is a hiking shoe (orange dotted box) which is a hard-to-classify positive. 
We want to train a generator that can learn to generate such hard samples.}
\label{fig:advgen}
\vskip -0.2in
\end{figure}

In this paper, we consider the problem of identifying query-item mismatches to enhance the ranking performance in product search. This task typically requires a large labeled dataset of matches and mismatches that we will respectively refer to as negative and positive samples. Even if we can partly afford the expensive and time-consuming labeling, acquired datasets are unbalanced and lack hard positive samples, preventing the classifier from learning a robust decision boundary. However, the above examples \texttt{gray Iphone X} and \texttt{Iphone X charger} motivate that meaningful positive samples can be artificially generated by leveraging the labeled data. In fact, we can heuristically construct a large number of negatives by observing which items are commonly purchased in response to the corresponding query. The question is that can we use such negatives to synthesize hard-to-classify positives to robustify the classifier? We illustrate the goal of the generation in Figure \ref{fig:advgen}.
 
To this end, we develop a deep, end-to-end model that learns to identify mismatched query-item pairs and is also capable of generating mismatched queries given an item. The task of the generator is twofold: it has to be able to generate hard-to-classify samples so that the classifier learns a more robust decision boundary; it also needs to generate realistic queries. 
 Using matched query-item pairs allows the generator to synthesize hard-to-classify mismatches based on an efficient encoder-decoder architecture. This has a distinct advantage over generating samples from noise, as in Generative Adversarial Networks \cite{goodfellow2014generative, wang2017irgan} or via dithering the learned representations to make the model more robust \cite{sngan}.  

We include our classifier and generator in an end-to-end model. 
The classifier only requires continuous representations of the generated query as the second input instead of a discrete text sequence. This key property enables us to use efficient gradient-based optimization techniques and bypass reinforcement learning-based methods \cite{jia2017adversarial}, which are significantly more complex, and also recently developed heuristic approaches to generate adversarial text samples \cite{alzantot-etal-2018-generating}. To achieve this, we modify the objective function in a way that makes the end-to-end training possible via sampling a binary latent variable, avoiding the min-max optimization for GANs \cite{sngan, wang2017irgan}. 

We perform extensive experiments on a mismatch dataset in an e-commerce company. The proposed model outperforms deep learning baselines by over $26 \%$ in F-score and $17 \%$ in relative AUPR score and performs significantly better than GBDT models, which are widely used in practice. Including the query generator helps achieve higher gains than merely dithering the vector representation of the query. We also show that the generative model can indeed generate hard-to-classify mismatches. When integrated with the ranking component of a real-world product search engine, our model outperforms the baseline methods in multiple countries on an online A/B test evaluation.

\subsection{Problem Setup}
\label{sec:setup}
Let $x = (I,Q)$ denote a pair of item title and textual query and $y(I,Q)$ denote its corresponding label. $y = 1$ if the pair is mismatched or $y = 0$ otherwise. 
Assume  we can obtain from search logs 
many matched samples, 
which we use to generate more positives. These samples are not human-labeled but instead inferred by considering behavioral signals such as frequent purchases. 

We aim to build a deep classifier that takes two text sequences in $x_i =(I_i, Q_i)$ and classifies whether the pair is mismatched or not. At the same time, we want the model to generate a new sample $(I, Q_{\text{gen}})$ with $y_{\text{gen}}=1$ given $(I, Q)$ with $y=0$. Next, we discuss our proposed model.  

\section{Proposed Model: QUARTS}
\label{sec:model}

\begin{figure}[h]
	\centering
	\includegraphics[width = 70mm, height=45mm]{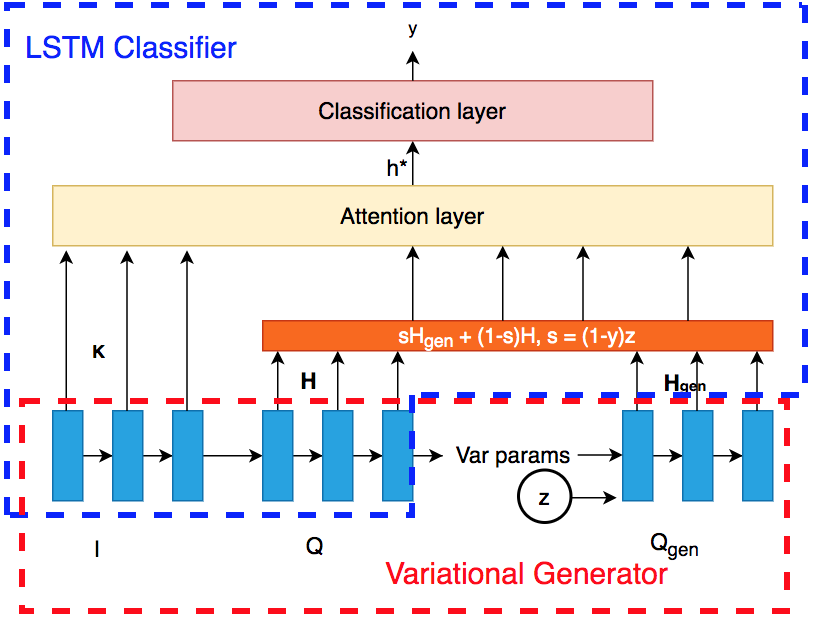} 
\caption{\small Our model (best seen in color). The blue dotted line encompasses the classifier. The red dotted line encompasses the generator. The orange layer in the model helps combine the outputs from the variational model and the original classifier. }
\label{fig:model}
\vskip -0.15in
\end{figure}

We present our proposed 
model, namely QUARTS (QUery-based Adversarial learning for Robust Textual Search) in Figure~\ref{fig:model}. QUARTS is composed of three components: (i) an LSTM and attention-based classifier,  (ii) a variational  encoder-decoder query generator (VED)  and (iii) a state combiner.

Due to space constraints, we defer the details of (i) and (ii) in the appendix. The LSTM classifier (i) is adapted from the entailment model in \cite{rocktaschel15_entailment}, with some changes to fit the product search task 
(see Appendix \ref{app:lstm}). 
The VED generator (ii) takes a matched  pair $(I, Q)$ as input and outputs a new query $Q_{\text{gen}}$ so that the pair $(I, Q_{\text{gen}})$ is mismatched while $Q_{\text{gen}}$ stays lexically similar to $Q$. As an example, if $I = $ \texttt{Apple Iphone X, space gray} and $Q = $ \texttt{gray Iphone X} is a matched pair, we can generate $Q_{\text{gen}} = $ \texttt{Iphone X case} given $I$. In this case, $Q_{\text{gen}}$ is similar to $Q$, but $(I, Q_\text{gen})$ constitutes a product mismatch.


To have an end-to-end model, we combine the query representations computed by the classifier and the generator to form a proper input to the attention layer. We need to make sure that the modifications still allow us to efficiently backpropagate the gradients of the loss function during training. To achieve this, we add a merging layer shown by the orange box in Figure \ref{fig:model}. This layer computes $sH_{\text{gen}} + (1-s)H, ~\ s = (1-y(I, Q))z$ where $H, H_{\text{gen}}$ are the corresponding LSTM representations of the input $Q$ and $Q_{\text{gen}}$, and $z\sim \textrm{Bernoulli}(p)$ is a random binary variable that controls whether the input query $Q$ or the generated query $Q_{\text{gen}}$ is used. When $z = 0$, QUARTS essentially computes the probability of mismatch.

Let us explain how the real label $y$ and the switch $z$ combine to yield the desired outputs. 
As $y=1$ where the sample $(I, Q)$ is a real positive, we want to leverage it to train the classifier $f_\theta(\cdot)$. In this case, $s = 0$ and the attention layer only takes $H$ as input. When $y=0$, we can either use this sample to train the classifier or use it to generate adversarial representations $H_\text{gen}$. This process is controlled by $z$. When $z = 1$, we use $H_\text{gen}$, else $H$. 
The value of $z$ determines whether we want to use the datapoint as-is for training, or instead use the \emph{``fake''} query via the VED module. 

A second consideration is how to enable efficient training  
on $f_\theta(\cdot)$ and the generator $g_\psi(\cdot)$. Let $x_\text{gen}=(I, Q_\text{gen})$ be the datapoint we will use to train $f_\theta(\cdot)$ using the output from $g_\psi(\cdot)$. In this case, since $y=0, ~\ z=1$, we use $z$ as a proxy ``label'' to train $f_\theta(\cdot)$. For samples $i=1, 2, \dots, N$, we sample $z_i \sim \textrm{Bernoulli}(p)$ for some $p \in [0, 1)$ to  decide which negative samples have labels flipped. We modify the cross entropy loss as below, with $L_\theta$ being the weighted cross-entropy loss:
\begin{equation}
\label{modloss}
\frac{1}{N} \sum_{i=1}^N (1- s_i)  L_\theta(x_i, y_i)  + s_i L_\theta(g_\psi(x_i), z_i)).
\end{equation}
Note that \eqref{modloss} is differentiable in $\theta, \psi$ and notably $H_\text{gen}$ -- the generated representations of $Q_\text{gen}$. Since we do not use the actual generated query, we need not resort to heuristics or policy gradient-based optimization methods to minimize \eqref{modloss}. Before training QUARTS end-to-end, we pre-train the classifier and the VED on proper data. The pseudocode of the end-to-end training is shown in Algorithm \ref{alg:e2e}.

\begin{algorithm}[!h]
  \caption{QUARTS training procedure}
  \label{alg:e2e}
  \begin{algorithmic}[1]
  \Require  $N$ samples of labeled data $(I, Q, y(I, Q))$, M \emph{negative} samples from search log, and sampling probability $p$ 
  \State Using labeled data, pre-train the classifier 
  \State Create $(I, Q_{\text{}}, Q_{\text{mis}})$ tuples $\mathcal{T}$ using labeled data so that $y(I, Q_{\text{}})=0$ and $y(I, Q_{\text{mis}})=1$
  \State Initialize the VED encoder with the trained classifier, and use the above created tuples to pre-train the VED generator 
  \State Concatenate the human annotated and logs data to form $M+N$ samples $\mathcal{D}$
  \State Perform end to end training on $\mathcal{D}$ , where in each epoch
  \For{$i \in [M+N]$}
  \State Sample $z \sim \textrm{Bernoulli}(p)$
  \State Set $s = (1 - y_i(I, Q))z$
  \State Use $s$ and $I, Q, y(I, Q)$ to perform one step of learning on the end-to-end model
  \EndFor
\end{algorithmic}
\end{algorithm}

\section{Experiments and Results}
\label{sec:exp}

We used a human-labeled dataset of query-item pairs, obtained from an e-commerce search platform. There are in total $N=3.2M$ pairs of which only a small fraction are mismatches. A separate test set of $ \sim100K$ labeled pairs was used to evaluate all methods. We further have $~3M$ query-item pairs that are deemed ``matched" by considering items that are purchased frequently in response to those queries from the search logs. This acts as the augmentation dataset for the QUARTS model. 

\subsection{Training Details}
\label{app:training}
For all encoders and decoders, we use an LSTM with hidden size of 300. The inputs to the encoder are 300 dimensional word embeddings trained separately for queries and item titles. The word embeddings were trained using word2vec  on a corpus of anonymized search engine queries, as well as item titles from the catalog. The models were trained using Adam ~\cite{kingma2014adam} and we tuned the classification part (i.e. excluding the variational decoder) on a validation dataset. We obtained the performance with initial learning rate $10^{-4}$, and learning rate decay $0.8$ after $10$ epochs. The drop-out probability and the batch size were respectively $0.1$ and $128$. Because the imbalanced nature of the labeled data, we up-weighted the positive samples. In the cross-entropy loss for classification, we set $\beta = 5$. 

To pretrain the VED, we used the annotated training data and generated $I, Q, Q_{\text{gen}}$ tuples as explained in Section \ref{app:ved}. Since we are explicitly interested in training the VED to generate $Q_{\text{gen}} : ~\ y(I, Q_{\text{gen}}) = 1$ given $I, Q: ~\ y(I, Q) = 0$, we consider only the annotated items that have both positive and negatively annotated queries, and generate the tuples. The previously pretrained encoder was fixed, and only the decoder was trained using Adam with an initial learning rate of $10^{-3}$. We finally merged the LSTM encoder for query and item, the VED decoder for query with the other layers described in the previous sections to train the model end to end.

The classifier $f_\theta(\cdot)$ is pretrained on the human annotated data. For the end-to-end model, we use the pretrained classifier and generator, modify the loss function as in \eqref{modloss}, and further append the dataset with $M=3MM$ well matched items from anonymized user logs, where we assume items that were purchased in response to a query are ``matched" $(y(I, Q) = 0)$.

\subsection{Metrics and Baselines} We evaluated our models using Area under the Precision-Recall curve (APR), and the F1-score at the best operating point, all evaluated on the test set. 
To evaluate the generation task, we used BLEU scores. In addition, we had human annotators to judge generated item-query pairs. These annotators were trained to identify whether a generated pair is a match or a mismatch. 

We used a GBDT model as a baseline. We used user-item features for this model similarly to traditional ranking and relevance models. 
We also applied a DSSM-style model (namely DSSM) where query and item word embeddings were concatenated as input to a stack of dense layers. 
We also used the BERT \cite{devlin2018bert} embeddings for the query and item title sequences and passed them through the aforementioned model. A final baseline we evaluated against was the MatchPyramid \cite{pang2016text}, which has shown to outperform several baselines for matching and question-answering tasks. All hyperparameters were chosen via a simple grid search on a validation set. All the results are reported on the test set. 



\subsection{QUARTS Performance}
The classification results of all considered models are shown in Table~\ref{tbl:score}. We also compare our model trained on the original training data and one augmented by naively adding the $3M$ matched pairs. For confidentiality reasons, we report the performance relative to some baseline. We see from Table \ref{tbl:score} that purely augmenting the training data with the matched samples does not improve but worsens the base classifier. 
Table \ref{tbl:scoredssm} shows the performance of the QUARTS compared with MatchPyramid models and the DSSM model initialized with pretrained BERT embeddings. The end-to-end QUARTS model beats the BERT DSSM baseline by over $17 \%$ in APR, and over $26 \%$ in F-score. 

\begin{table}[h!tb]
  \centering
  \small
\begin{tabular}{lllll}
\hline
Model	& APR 	& F-score		\\
\hline
GBDT 					& baseline 	& baseline \\
DSSM 					& +26.16$\%$ 	& +28.86$\%$  \\
DSSM $+$ BERT 			& +33.71$\%$  	& +37.56$\%$  \\
MatchPyramid 				& +44.95$\%$ 	& +40.09$\%$ \\
QUARTS Classifier 			& +52.06  $\%$	& +55.21$\%$  \\
QUARTS Classifier + Augment & +50.9$\%$	& +51.5$\%$  \\
QUARTS end-to-end 		& +56.65$\%$ 	& +62.43$\%$\\
\hline
\end{tabular}
\caption{\small The classification performance of our model on average precision and F-score, compared with baselines. The performance is relative to a GBDT model. }
\label{tbl:score}
\vskip -0.15in
\end{table}


\begin{table}[!h]
  \centering
\small
\begin{tabular}{lllll}
\hline
Model	& APR 	& F-score		\\
\hline
DSSM $+$ BERT 			& baseline  	& baseline  \\
MatchPyramid 				& +8.4$\%$ 	& +16.44$\%$ \\
QUARTS Classifier 			& +13.72$\%$	& +20.85$\%$  \\
QUARTS end-to-end 		& +17.15$\%$ 	& +26.06$\%$\\
\hline
\end{tabular}
\caption{\small Comparison with other deep learning baselines. 
}
\label{tbl:scoredssm}
\vskip -0.15in
\end{table}
To validate the effectiveness of QUARTS in improving the ranking performance for the search task, we performed an A/B test on live search traffic in two countries, to account for varying traffic patterns.  
Compared to the existing baselines, the QUARTS model yielded a $ 12.2 \%$  and $ 5.75\%$ increase in online metrics for the two countries respectively, which are significant given the task.

\begin{table*}[t]
\small
\begin{tabular}{lll}
\hline
Item title $(I)$ & Query $(Q)$ & Generated query $(Q_{\text{gen}})$ \\
\hline
  {\small ESR iPhone 8/7 screen protector tempered glass... } & iPhone 8 curved \textbf{screen protector}	& iPhone 8 plus \textbf{cases} \\
{\small JETech case for iPad Pro 12.9 inch} & ipad pro 12.9 speck \textbf{shell}	& \textbf{iPad pro} 12.9 \\ 
{\small Mounting dream full motion  wall mounts bracket } & lg oled \textbf{tv mount}	 & 55 inch \textbf{flat screen tv} \\ 
{\small Intel core i7-8700K desktop processor 6 cores} & \textbf{core i7 8700k}	 & \textbf{GTX 1080} \\ 
{\small Chicco pocket snack booster seat} & peg perego \textbf{high chair} & baby \textbf{dining set} \\ 
{\small Comfy sheets ultra luxury 100\% Egyptian cotton sheet set} & king size \textbf{sheets}	 & king size \textbf{beds} for sale \\ 
\hline
\end{tabular}
\caption{\small Examples of adversarial query generations from the VED query generator. The Item and Query should be matched, while the Item and generated query should be mismatched. For readability, we have \textbf{bolded} words in the query and generated query to show how the VED changes the product type intent in the generated query, while still being similar to the original query. }
\label{tbl:vedqual}
\vskip -0.15in
\end{table*}

\subsection{VED Results}
We used a held-out $10\%$ of the $(I, Q, Q_{\text{gen}})$ data to evaluate the VED generator. In order to make a fair evaluation, we ensured that the items that appeared in training set were not in the validation set. The validation BLEU scores are shown in Table~\ref{tbl:bleu}. BLEU scores do not indicate whether or not a generated queries is a ``realistic" modification of the original query. 
Therefore, we also had 2500 generated pairs annotated by human experts who were specifically trained to decide if a query-item pair is matched or not. The accuracy $82\%$ in Table \ref{tbl:bleu} suggests that most of generated pairs are meaningful. Here, the accuracy is the fraction of the pairs that were actually labeled as mismatches 

\begin{table}[h!tb]
  \small
  \centering
\begin{tabular}{|l|llll|l||}
\hline
\small Model	& \scriptsize BLEU-1	& \scriptsize BLEU-2	& \scriptsize BLEU-3	& \scriptsize BLEU-4  & \small Acc\\
\hline
VED &35.15 & 31.40 & 24.84 & 20.76 & 0.82 \\
\hline
\end{tabular}
\caption{\small Validation BLEU scores of generated queries from the variational encoder-decoder generator, and misclassification accuracy as reported by humans.}
\label{tbl:bleu}
\vskip -0.1in
\end{table}

We provide some qualitative results from the VED in Table \ref{tbl:vedqual}. The generator's goal is to slightly modify the input query $Q$, so that the resultant $(I, Q_{\text{gen}})$ sample is realistic. 
A source query for \texttt{screen protector} is mapped to a query for \texttt{phone case}, and a source query for \texttt{tv mount} is mapped to one for \texttt{flat screen tv}.

\subsection{Word-by-Word Attention Visualization}
\begin{figure}[!h]
\centering
\includegraphics[width = 55mm, height = 25mm]{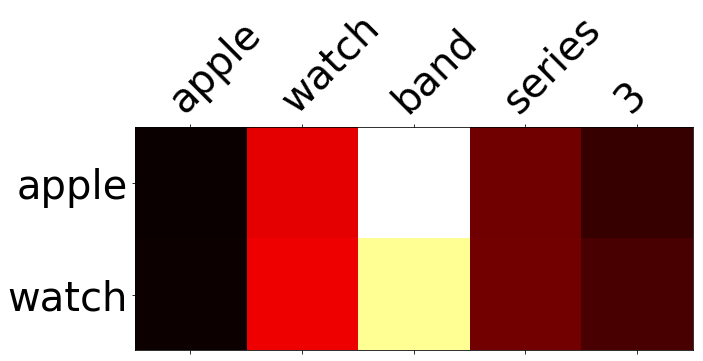} 
\includegraphics[width = 55mm, height = 25mm]{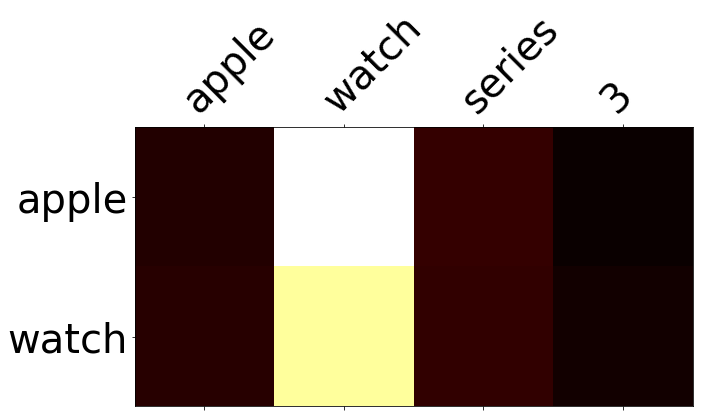} 
\caption{\small Word-by-word attention for a mismatched (top) and matched (bottom) query-item pair. Rows represent query words, columns represent item words, with lighter shares representing larger weights.  \texttt{band} is attended to more on the left whereas \texttt{watch} is attended to more in the right }
\label{fig:wbwatt}
\vskip -0.15in
\end{figure}

\begin{figure}[!h]
\centering
\includegraphics[width = 55mm, height = 28mm]{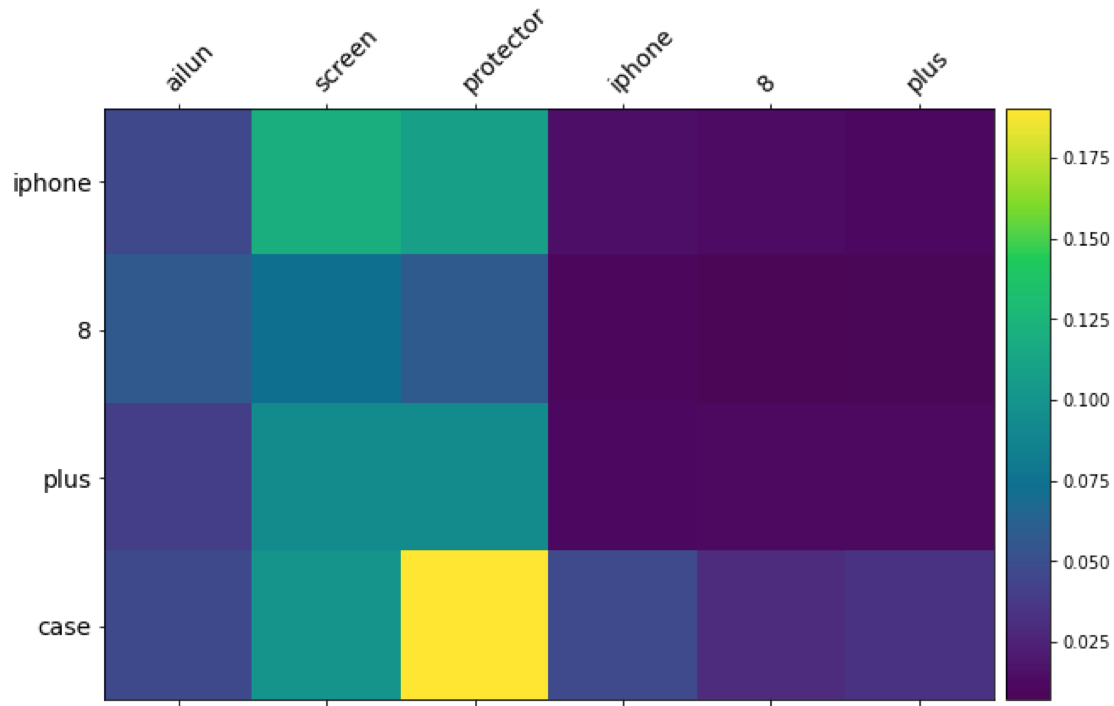} 
\includegraphics[width = 55mm, height = 28mm]{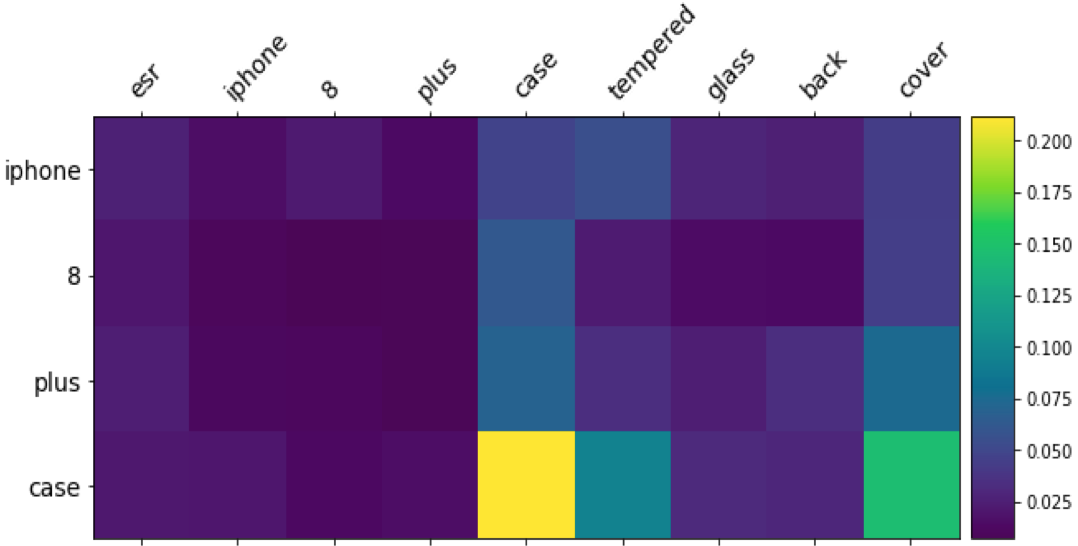} 
\caption{ \small Word-by-word attention for a mismatched (top) and matched (bottom) query-item pair. Rows represent query words, columns represent item words, with lighter shares representing larger weights.  \texttt{protector} is attended to more on the left whereas \texttt{case} is attended to more in the right }
\label{fig:wbwattcase}
\vskip -0.15in
\end{figure}

The goal of the word-by-word attention layer is to understand what parts of the user query and item titles are important to understand whether to match or not.  Importantly,  item titles are typically long, and have information such as brand, color and size. All of these facets might not be relevant for a particular user query. Figure \ref{fig:wbwatt} shows the performance of the word-by-word attention layer, for a matched and a mismatched pair. In both cases, we see that the correct words are attended to, helping the classifier make the distinction between a matched and a mismatched pair. Figure \ref{fig:wbwattcase} shows another example. 


\section{Conclusion and Future Work}
\label{sec:conc}
We developed an end-to-end  model with hard to classify query generation for retrieval in e-commerce product search. We built upon ideas for textual entailment, and used a word by word attention layer to help create item representations conditioned on an input query. We trained a generator that yields representations of queries that are mismatched to a source item, while at the same time being ``realistic".  This allows us to address the class imbalance of our datasets, while also generating samples that help robustly train the classifier. To train the model end to end, we modified the cross-entropy loss, allowing us to avoid optimizing a minimax objective. Experiments on an offline dataset and live product search traffic showed that our method improves significantly over baselines.

\bibliography{refs}
\bibliographystyle{acl_natbib}

\onecolumn
\appendix

\section{Appendices}
\label{sec:appendix}

\subsection{LSTM Classifier}
\label{app:lstm}

\begin{figure}[h]
	\centering
	\includegraphics[scale=.5]{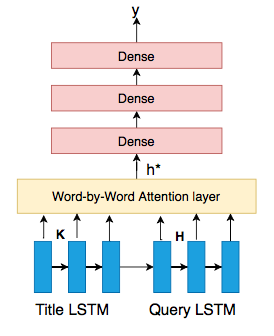} 
\caption{LSTM based classifier for a query-item pair (best seen in color). The LSTMs are fed word embeddings, separately learned for queries and titles. The word-by-word attention layer is adapted from ~\cite{rocktaschel15_entailment}, and $h^*$ is defined in \eqref{clfinput}}
\label{fig:clf}
\vskip -0.15in
\end{figure}

We adapt our classifier from that for textual entailment in~\cite{rocktaschel15_entailment}, but with a few key differences. Unlike standard textual entailment problems for natural language, user queries and item titles tend to follow different language patterns, with both of them being different from ``natural" language. For example, queries \texttt{"red nike running shoes"}, \texttt{"running nike shoes, red"} and \texttt{"red running shoes nike"} all refer to the same general product, despite differing in structure. On the other hand, item titles are structured, with brand, size, color, etc. all mentioned in a long sequence, which is also not how a conventional sentence is structured. To account for these differences between query strings and item titles, we separately train word embeddings using word2vec \cite{mikolov2013distributed} on anonymized query logs and item titles. Thus, the same word can have two embeddings, one for the query and one for the title. The overall classifier structure is shown in Figure \ref{fig:clf}

We implement the word-by-word attention layer as follows: Let $k$ be the output dimension of the LSTMs, $K \in \R^{k \times m}$ and $H = [h_1, h_2, \dots, h_n] \in \R^{k \times n}$ be the LSTM output matrices for the item title and query respectively, with the $i^{th}$ column corresponding to the output of the $i^{th}$ LSTM cell. Let $m$ and $n$ denote respective lengths of title and query sequences. 

For each word $t=1, 2, \dots, n$ in the query, we compute attention scores for every word in the title and its weighted representation $r_t$ at that step. The representation $r_{t-1}$ is helpful to inform the next step what the model previously paid more focus on. We use the additive attention ~\cite{bahdanau2014_attention} here, but other alternatives can be used as well.
\begin{align*}
& M_t = \tanh([K^\top, \1h_t ^\top, \1 r_{t-1}^\top]W_h), \quad W_h \in \R^{3k\times k} & \\
& \alpha_t = \tanh(M_t w), \quad w \in \R^{k} & \\
& r_t =  K^\top\alpha_t + \tanh(W_r r_{t-1} ), \quad W_r \in \R^{k \times k}. &
\end{align*}

The final representation for the query and title that is passed to fully-connected layers is:
\begin{equation}
\label{clfinput}
h^* = \tanh(W_x [r_n^\top, q_n^\top, |r_n - q_n|^\top]). ~\ W_x \in R^{k\times 3k}.
\end{equation}
In the above equations, $W_h, w, W_r$ and $W_x$ are weight matrices to be learned, and $q_n$ is the output of the LSTM that encodes the query. Passing $|r_n - q_n|$ in \eqref{clfinput} to the dense layers improves classification performance. We observe in mismatched query-item pairs that a slight word substitution or deletion often leads to mismatched items; for example, \texttt{"iPhone screen protector"} and \texttt{"iPhone screen"} or \texttt{"iPad screen protector"} are textually very similar, but are completely different items from a shopping point of view. Hence, we use the term $|r_n - q_n|$ in \eqref{clfinput} to explicitly account for such word changes. Traditional sentence classification methods also pass $r_n \circ q_n$ to the dense layers, where $\circ$ is the hadamard product. We noticed that this did not improve the model performance, and hence choose to not use it. A desirable side effect is reduced computations. We expect that $|r_n - q_n|$ somehow captures words that are in the query but not in the title and vice versa.

Let $f_\theta(\cdot)$ denote the classifier in Figure \ref{fig:clf}. Given $N$ samples $\{(x_i, y_i)\}_{i=1}^N$, our objective function is a weighted cross-entropy loss:
\begin{align}
\label{lf}
&L_\theta(X, y) = \frac{1}{N}\sum_{i=1}^N L_\theta(x_i, y_i) \\
\notag
&= \frac{1}{N}\sum_{i}\beta  y_i \log(f_\theta(x_i))  +  (1-y_i) \log(1 - f_\theta(x_i))
\end{align}
where $\beta$ adjusts the weight on the positive samples. We set $\beta > 1$  to account for the fact that the number of positive samples (i.e. mismatched) is much larger than negative samples in our datasets.  

\subsection{Variational Query Generator}
\label{app:ved}

For the applications we are interested in, the training datasets are highly unbalanced, as a reasonable search engine will have far more matched query-item examples than mismatched. We thus need ways to account for this class imbalance. Generating trivially mismatched examples is easy: we can randomly sample an item from the entire catalog for a given query.  But these will be examples that are easy-to-classify for $f_\theta(\cdot)$, and are hence uninformative. Here we aim to train a model that can generate hard-to-classify mismatched examples, which tend to occur due to the query and product title being lexically similar. Specifically, we want to generate mismatched query-item examples that have a realistic chance of appearing in the search results for said query.  

We train a Variational Encoder-Decoder (VED) model to this end. The model takes as input a matched  pair $(I, Q)$, and outputs a new query $Q_{\text{gen}}$ so that the  pair $(I, Q_{\text{gen}})$ is mismatched, but being lexically similar to $Q$. As an example, if $I = $ \texttt{puma running shoe, size 11, black} and $Q = $ \texttt{running shoes for men}, we can generate $Q_{\text{gen}} = $ \texttt{insoles for running shoes}. In this case, $Q_{\text{gen}}$ is similar to $Q$ in that the item is somewhat related to $Q_{\text{gen}}$, and there's a chance that $I$ may be matched to $Q_{\text{gen}}$ due to keyword stuffing by sellers, or poor semantic matching. On the other hand, another mismatched query $Q_{\text{gen}} = $ \texttt{pizza cutter} is not a good candidate to generate, since it's highly unlikely that a reasonable search engine will show shoes for a query about pizza cutters. 

\begin{figure}[h]
	\centering
	\includegraphics[width = 70mm, height=30mm]{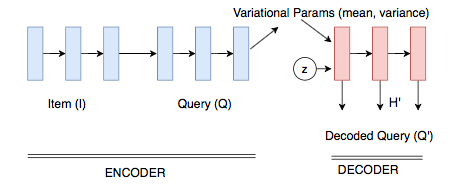} 
\caption{Variational encoder-decoder query generator (best seen in color). The encoder is reused from the classifier in the previous section. The decoder is an LSTM with attention ~\cite{luong2015_attention}. $Q_{\text{gen}}$ is a generated query via beam search. }
\label{fig:ved}
\end{figure}

To train the model, we make use of an labeled $\{(I, Q, y)\}$ dataset and create a new one as follows: we consider only those items $I$, for which there exist both matched and mismatched queries, and construct samples $(I, Q_{\text{}}, Q_{\text{mis}})$ so that $y(I, Q_{\text{}})=0$, and $y(I, Q_{\text{mis}}) = 1$. The model is the variational sequence to sequence model proposed in \cite{ved}, which we adapt to our case (Figure \ref{fig:ved}). Our architecture can reuse the title-query encoder of the classifier in Section \ref{app:lstm}. The variational decoder allows us to generate diverse output sequences for the same input. We equip the decoder with an attention mechanism~\cite{luong2015_attention} to generate $Q_{\text{gen}}$. Using an existing annotated dataset to pretrain the VED allows us to accurately warm start the end to end model described in the next section. 

\subsection{Learned LSTM Embeddings}
\label{app:embeddings}
Next, we verify that the learned query and item embeddings from the LSTM models are informative. To compute a vector representation of a query, we mean-pool the query LSTM outputs $H$. Similarly, for an item, we mean-pool the item LSTM outputs $K$. Table \ref{tbl:knnq} shows the 3-nearest neighbors for user queries. The neighbors are computed based on the cosine similarity between the embeddings of the source and target query. We can see that, depending on the specific query, the model learns to group queries that have the same product intent, or brand (for higher-end) items. 

\begin{table*}[!h]
  \small
  \centering
\begin{tabular}{| l | l |}
\hline
source &\textbf{screen replacement for iphone 7 plus in white including all tools instruction 2 screen protectors repair kit with digitizer} 	\\
target 1& {iphone 7 plus screen replacement white lcd display 3d touch screen digitizer frame assembly white} \\
target 2&		 {iphone 6 screen replacement white p zone 4 7 inch lcd display touch screen digitizer frame assembly} \\
target 3& {for iphone 7 screen replacement lcd touch screen digitizer frame assembly full set} \\
\hline
source&\textbf{imagine by rubie s dc superheroes harley quinn mallet costume} 	\\
target 1& {rubie s harley quinn mallet costume accessory} \\
target 2&		 {rubie s women s suicide squad harley quinn mallet as as shown one size} \\
target 3&		 {rubie s women s batman harley quinn inflatable mallet multi one size} \\
\hline
source&\textbf{nike unisex core golf visor dark grey anthracite white one size} 	\\
target 1& {nike golf unisex legacy91 hat white black one size} \\
target 2&		 {nike men s flex core golf shorts dark grey dark grey size 36} \\
target 3&		 {nike golf tech visor black adjustable one size} \\
\hline
source&\textbf{mercer culinary genesis 6 piece forged knife block set tempered glass block} 	\\
target 1&{dalstrong knife set block gladiator series knife set german hc steel 8 pc} \\
target 2&		 {j a henckels international 13550 005 statement knife block set 15 pc light brown} \\
target 3&		 {top chef by master cutlery 5 piece chef basic knife set with nylon carrying case} \\
\hline
source&\textbf{hicksholsters purple dark punisher edition wallet} 	\\
target 1& {hicksholsters kydex dark punisher edition wallet} \\
target 2&		 {silk iphone 6 6s wallet case vault protective credit card grip cover wallet slayer vol 1 black onyx} \\
target 3&		 {kalmore genuine leather rfid protected slim thin pocket wallet minimalist wallet money clip light blue with id window} \\
				
\hline
\end{tabular}
\caption{\small Three nearest neighbors by cosine similarity for a few items. In each cell, the first line (bolded) represents the source, and the next 3 lines represents the 3 nearest neighbors. Unlike the query case, the nearest neighbors are always substitutable items. }
\label{tbl:knni}
\end{table*}

\begin{table}[!h]
  \small
  \centering
\begin{tabular}{|l|l|}
\hline
Source	& Targets 		\\
\hline
{kate spade yoga mat} 	& {kate spade} \\
		& {kate spade sale} \\
		& {kate spade wallet} \\
\hline
{dickies overalls striped} 	& {dickies work pant} \\
		& {32x32 mens dickies shorts} \\
		& {kahki overalls for women} \\
\hline
{puppy training 101} 	& {dog training pad} \\
		& {dog training collar} \\
		& {puppy} \\
\hline
{plastic stacking bins} 	& {stackable storage bins} \\
		& {storage bins} \\
		& {foldable storage bins} \\
				
\hline
\end{tabular}
\caption{\small Three nearest neighbors by cosine similarity for a few queries.  Note that the LSTM + mean-pooling method accurately represents queries based on various intents. In the first case, customers looking for Kate Spade items tend to look for more items of the same brand. In the second case, the model groups queries with similar intents together. In the last 2 cases, the model groups queries that are substitutable together. }
\label{tbl:knnq}
\end{table}

Along similar lines, Table \ref{tbl:knni} shows the 3-nearest neighbors in terms of cosine similarity for items. Note that this case is not the same as queries, since an item by itself is meaningless. Indeed, the outcome of the word by word attention model is to achieve an item representation conditioned on the query. More specifically, the item \texttt{nike running shoe} by itself cannot be deemed as matched or mismatched, unless seen in the context of a user typed query. Hence, the LSTM + mean pooling output for the items tend to cluster similar (substitutable) items together. The upshot if this is, conditioned on a given query $Q$, the item embeddings for similar items will be similar, which is a desirable outcome for our use case. 

\subsection{Related Work}
\label{sec:related}
The DSSM \cite{huang2013learning} model and it's variants  \cite{mitra2017learning, xiong2017end} have been commonly applied in learning to  rank tasks. Such models are useful for web search, where there are several related documents and it's easier for natural language based models to distinguish between related and unrelated documents. These models do not easily carry over for product search, due to the issues alluded to in the previous section. Recently, \cite{kang2018adventure} developed means to generate adversarial samples to improve entailment, via the use of additional datasets to learn ``rules" to aid in sample generation. These rules do not carry over to the product search domain, nor do the assumption of existing datasets to learn such rules. To the best of our knowledge, we are the first to work on generating adversarial representations of text for the purpose of improving product relevance for e-commerce.

Adversarial example generation has been studied in the context of images \cite{szegedy2013intriguing, chen2018ead}, speech \cite{carlini2018audio} and text \cite{cheng2018seq2sick, ebrahimi2017hotflip, kuleshov2018adversarial, iyyer2018adversarial, papernot2016crafting, kang2018adventure, wang2017irgan}. In \cite{alzantot-etal-2018-generating, ebrahimi2017hotflip}, the authors develop a means to perturb the sequence in order to fool an underlying classifier, and in \cite{iyyer2018adversarial}, the authors use the concept of back-translation \cite{sennrich2015improving}. The aims in these works is to generate adversarial text samples themselves, separate from generating samples that will make the underlying classifier more robust.  

\end{document}